\documentclass[12pt]{article}

\usepackage{verbatim}
\usepackage{float}
\usepackage{pgfplots}
\usepackage{tikz}
\usepackage{amsmath}
\usepackage{subfigure}
\usepackage{multirow}
\usepackage{booktabs}
\usepackage[margin=1in]{geometry}
\usepackage[noadjust]{cite}

\providecommand{\keywords}[1]{\textbf{Keywords---} #1}

\pgfplotsset{compat=1.16}

\usetikzlibrary{quotes,arrows.meta}
\tikzset{
  annotated cuboid/.pic={
    \tikzset{%
      every edge quotes/.append style={midway, auto},
      /cuboid/.cd,
      #1
    }
    \draw [every edge/.append style={pic actions, densely dashed, opacity=.5}, pic actions]
    (0,0,0) coordinate (o) -- ++(-\cubescale*\cubex,0,0) coordinate (a) -- ++(0,-\cubescale*\cubey,0) coordinate (b) edge coordinate [pos=1] (g) ++(0,0,-\cubescale*\cubez)  -- ++(\cubescale*\cubex,0,0) coordinate (c) -- cycle
    (o) -- ++(0,0,-\cubescale*\cubez) coordinate (d) -- ++(0,-\cubescale*\cubey,0) coordinate (e) edge (g) -- (c) -- cycle
    (o) -- (a) -- ++(0,0,-\cubescale*\cubez) coordinate (f) edge (g) -- (d) -- cycle;
  },
  /cuboid/.search also={/tikz},
  /cuboid/.cd,
  width/.store in=\cubex,
  height/.store in=\cubey,
  depth/.store in=\cubez,
  units/.store in=\cubeunits,
  scale/.store in=\cubescale,
  width=10,
  height=10,
  depth=10,
  units=cm,
  scale=.1,
}

\def\camera#1#2{
\begin{scope}[shift={#1}, rotate=#2]
    \draw [fill=black](0,0) -- (0.5,1) -- (-0.5,1) -- cycle;
    \draw [fill=white,ultra thick](0,0) circle (0.2 5);
\end{scope}
}

\date{}

\begin{document}

\title{Regression on Deep Visual Features using Artificial Neural Networks (ANNs) to Predict Hydraulic Blockage at Culverts}

\author{\small{Umair Iqbal\thanks{Umair Iqbal is with the SMART Infrastructure Facility, University of Wollongong, Australia, e-mail-ui010@uowmail.edu.au}, Johan Barthelemy, Pascal Perez and  Wanqing Li}% <-this % stops a space
}

\maketitle

\begin{abstract}
Cross drainage hydraulic structures (i.e., culverts, bridges) in urban landscapes are prone to getting blocked by transported debris which often results in causing the flash floods. In context of Australia, Wollongong City Council (WCC) blockage conduit policy is the only formal guideline to consider blockage in design process. However, many argue that this policy is based on the post floods visual inspections and hence can not be considered accurate representation of hydraulic blockage. As a result of this on-going debate, visual blockage and hydraulic blockage are considered two distinct terms with no established quantifiable relation among both. This paper attempts to relate both terms by proposing the use of deep visual features for prediction of hydraulic blockage at a given culvert. An end-to-end machine learning pipeline is propounded which takes an image of culvert as input, extract visual features using deep learning models (i.e., MobileNet, ResNet50, EfficientNetB3), pre-process the visual features and feed into regression model (i.e., Artificial Neural Network (ANN)) to predict the corresponding hydraulic blockage. Dataset (i.e., Hydrology-Lab Dataset (HD), Visual Hydrology-Lab Dataset (VHD)) used in this research was collected from in-lab experiments carried out using scaled physical models of culverts where multiple blockage scenarios were replicated at scale. Performance of regression models was assessed using standard evaluation metrics including Mean Squared Log Error (MSLE), Mean Squared Error (MSE), Mean Absolute Error (MAE) and $R^{2}$ score. Furthermore, performance of overall machine learning pipeline was assessed in terms of processing times for relative comparison of models and hardware requirement analysis. From the results ANN used with MobileNet extracted visual features achieved the best regression performance with $R^{2}$ score of 0.7855. Positive value of $R^{2}$ score indicated the presence of correlation between visual features and hydraulic blockage and suggested that both can be interrelated with each other. 
\end{abstract}


\section{Introduction}

Blockage of cross drainage hydraulic structures such as culverts and bridges is a commonly occurring phenomena during floods which often results in increased damages to property and risk to life \cite{french2015culvert, french2018design, BarthelmessMechanism, rigby2001impact, roso2004prediction, rigby2002causes, davis2001analysis, iqbalcomputer}. However, consideration of blockage into design guidelines is hindered by highly variable nature of blockage formulation and unavailability of historical during floods data to investigate the behaviour of blockage \cite{kramer2015physical, blanc2013analysis, blanc2014analysis}. In context of Australia, many councils and institutions have mentioned blockage as critical issue (e.g., NSW Floodplain Management Manual \cite{NSWfloodplain}, Queensland Urban Drainage Manual \cite{jones1991queensland}, Australian Rainfall and Runoff (ARR) \cite{french2018design, ollett2017australian, ARR_Report, ball2016australian}), however, none comprehensively addressed consideration of blockage into design guidelines. Wollongong City Council (WCC) under the umbrella of ARR developed a conduit blockage policy in 2002 which was revised many times over the years \cite{CondouitBlockagePolicy, ARR_Report}. This policy was developed based on the post flood visual surveys of hydraulic structures and suggested that any hydraulic structure with diagonal length less than 6m is prone to 100\% blockage during peak floods \cite{Barthelmess2009Quantification, rigby2001impact, rigby2002causes, CondouitBlockagePolicy}. 

However, WCC blockage policy was criticized by the hydrology experts because of its dependence on the visual assessments rather than the hydraulic assessments. Furthermore, it is argued that the ``degree of blockage" in WCC policy was assessed from the post flood visual inspections which cannot be considered as true representation of blockage during peak floods \cite{french2012non}. As a result of this on-going debate, visual blockage and hydraulic blockage are introduced as two separate terms. Visual blockage is defined as the percentage occlusion of hydraulic structure by debris assessed from images and/or personal observations. However, it is not necessary for a structure with high visual blockage to be hydraulically blocked as well (e.g., culvert blocked with porous vegetative debris have high visual blockage but may have insignificant effect on hydraulic capacity of culvert). Hydraulic blockage, on the other hand is defined as the reduction in the hydraulic capacity of structure due to presence of debris material \cite{ARR_Report}. Quantification of hydraulic blockage at a given structure is highly variable and almost impossible to achieve. French et al. \cite{french2012non} suggested that unless the visual blockage is translated into hydraulic blockage, the WCC policy cannot be considered reliable for including into design guidelines. Furthermore, hydraulic data from peak floods is needed to better understand the blockage behaviour and its impact on the floods. Till date, there is no quantifiable relation reported between visual blockage and hydraulic blockage. 

Research in this paper attempts to relate hydraulic blockage with visual blockage by proposing an end-to-end machine learning pipeline. Aim of this approach is to take the image of culvert and predict the corresponding hydraulic blockage. The proposed machine learning pipeline involves the steps of  extraction of visual features from image using deep learning models (i.e., MobileNet, ResNet50, EfficientNetB3), pre-processing of the extracted deep visual features and predicting the hydraulic blockage by feeding it to regression model (i.e., Artificial Neural Network (ANN)). The dataset (i.e., Hydrology-Lab Dataset (HD), Visual Hydrology-Lab Dataset (VHD)) used in this research was collected from series of comprehensive in-lab experiments performed using scaled physical models of culverts to replicate different flooding and blockage scenarios at scale. 

Rest of the paper is organized as follows: Section 2 presents the information about the visual and hydraulic dataset used in this research. Section 3 presents the information about proposed end-to-end machine learning pipeline and theoretical background of methods used in the pipeline. Section 4 provides the information about the model hyperparameters and evaluation measures used to assess the performance of regression models. Section 5 presents the results of investigation and highlights the important insights extracted from the results. Section 6 concludes the study by reporting important results, highlighted trends and potential future directions. 

\section{Dataset}

Two different types of datasets (i.e., hydraulic and visual) were used in this research collected from comprehensive in-lab experiments using scaled physical models of culverts. Aim of the experiments was to replicate blockage scenarios using multiple debris type during different flooding conditions at scale and record both visual and hydraulic data. Percentage hydraulic blockage was recorded using mathematical formulation proposed by Kramer et al. \cite{kramer2015physical} as given in Equation 1. 

\begin{equation}
    \text{Percentage Hydraulic Blockage} = \frac{\text{Upstream WL}_{\text{blocked}}-\text{Upstream WL}_{\text{unblocked}}}{\text{Upstream WL}_{\text{blocked}}}\times 100
\end{equation}

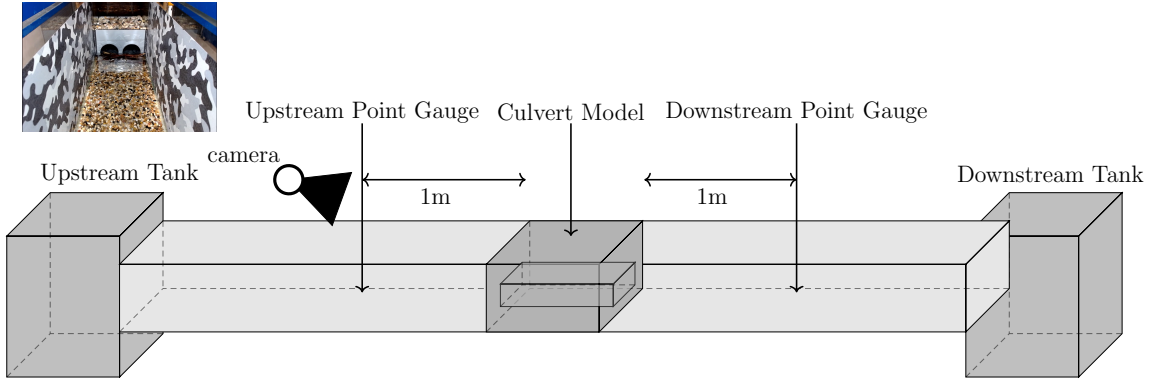
\begin{figure}
 \centering
 \scalebox{0.75}{\begin{tikzpicture}
%\draw[step=1cm,gray!20,very thin] (0,-10) grid (20,10);
\pic[fill=gray!50, text=black, draw=black] at (0,0) {annotated cuboid={width=20, height=25, depth=20}};
\pic[fill=gray!50, text=black, draw=black] at (17,0) {annotated cuboid={width=20, height=25, depth=20}};
\pic[fill=gray!20, text=black, draw=black] at (15,-0.5) {annotated cuboid={width=150, height=12, depth=20}};

\pic[fill=gray!60, text=black, draw=black] at (8.5,-0.5) {annotated cuboid={width=20, height=12, depth=20}};
\pic[fill=gray!60, text=black, draw=black, opacity=0.7] at (8.75,-0.85) {annotated cuboid={width=20, height=4, depth=10}};

\camera{(3,1)}{-110}

\node[] () at (0, 3){\includegraphics[scale=0.06]{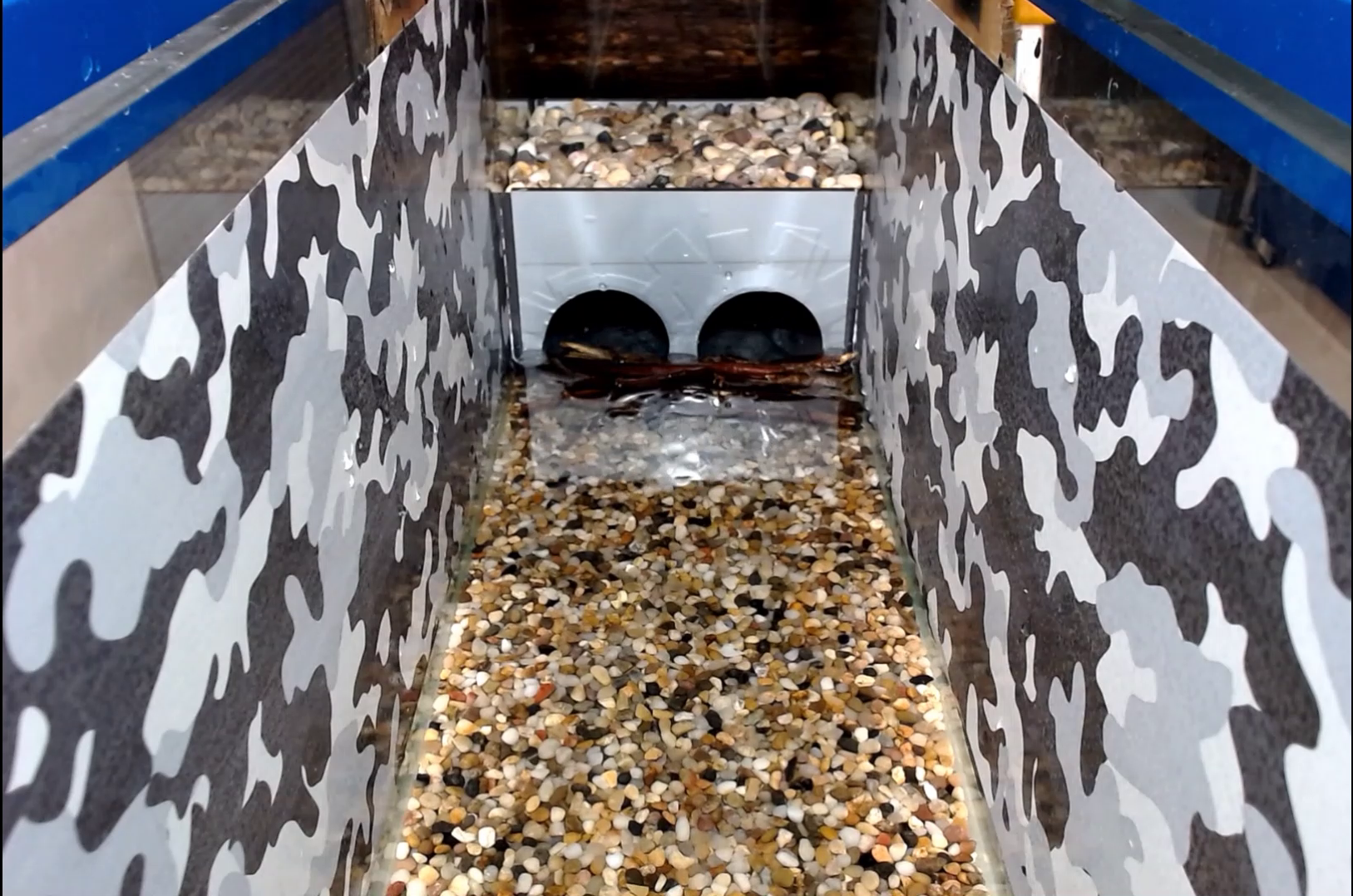}};

\node[align=left] at (2.2,1.4) {camera};
\node[] at (0,1.1) {Upstream Tank};
\node[] at (16.5,1.1) {Downstream Tank};

\node[] at (5.6,0.7) {1m};
\node[] at (10.5,0.7) {1m};

\node[] at (4.3,2.2) {Upstream Point Gauge};
\node[] at (12,2.2) {Downstream Point Gauge};
\node[] at (8,2.2) {Culvert Model};

\draw[thick, ->] (4.3,2) to (4.3,-1);
\draw[thick, ->] (12,2) to (12,-1);
\draw[thick, ->] (8,2) to (8,0);

\draw[thick, <->] (12,1) to (9.3,1);
\draw[thick, <->] (7.2,1) to (4.3,1);
\end{tikzpicture}}
 \caption{Two-Dimensional Schematic Diagram of Experimental Setup.}
 \label{fig:experimental_Setup}
\end{figure}

Experiments were performed in a 12m $\times$ 0.2m flume with single and double circular culvert models. Vegetative and urban debris was used at scale to simulate different blockage scenarios. Figure \ref{fig:experimental_Setup} shows the Two-Dimensional (2D) schematic diagram of the experimental setup used to collect the dataset. Point gauge was used to measure the water levels and was placed at 1m distance from culvert. In total, 173 unique blockage scenarios were simulated while some scenarios were repeated. Total of 352 hydraulic data samples were recorded from the experiments to organize in a dataset called Hydrology-Lab Dataset (HD).

In addition to collection of hydraulic data, a web camera-based setup was established to record the videos of each simulated blockage scenario. The visual dataset is referred as Visual Hydrology-Lab Dataset (VHD). For this investigation, images were extracted from VHD for the time instances when the hydraulic measurements were taken. In total 352 images were extracted from video clips each representing the visual of culvert for the time instance at which corresponding hydraulic measurement was taken.

\section{Methodology}

This section provides the information about the proposed end-to-end machine learning pipeline and theoretical background of methods used in the pipeline. 

\begin{figure}
 \centering
\scalebox{0.9}{\begin{tikzpicture}

%\draw[step=1cm,gray!20,very thin] (0,0) grid (16,10);
\tikzstyle{rectb}=[rectangle,draw=black, thick, minimum height=2cm, minimum width=3cm],
\tikzstyle{rectc}=[rectangle,draw=black, thick, minimum height=4cm, minimum width=7cm],

\node (input) at (0,8){\includegraphics[scale=0.05]{Figures/Sample_199.png}};
\node[rectb](feature) at (4,8){\begin{tabular}{c} CNN Feature \\ Extraction  \end{tabular}};
\node[rectb](processing) at (8,8){\begin{tabular}{c} Data \\ Processing \end{tabular}};
\node[rectb](Regression) at (12,8){\begin{tabular}{c} ANN \\ Regression \end{tabular}};
\node (out) at (15.5,8) {\begin{tabular}{c} \% Hydraulic \\ Blockage \end{tabular}};

\node[rectc](image1) at (4,4) {\includegraphics[scale=0.225]{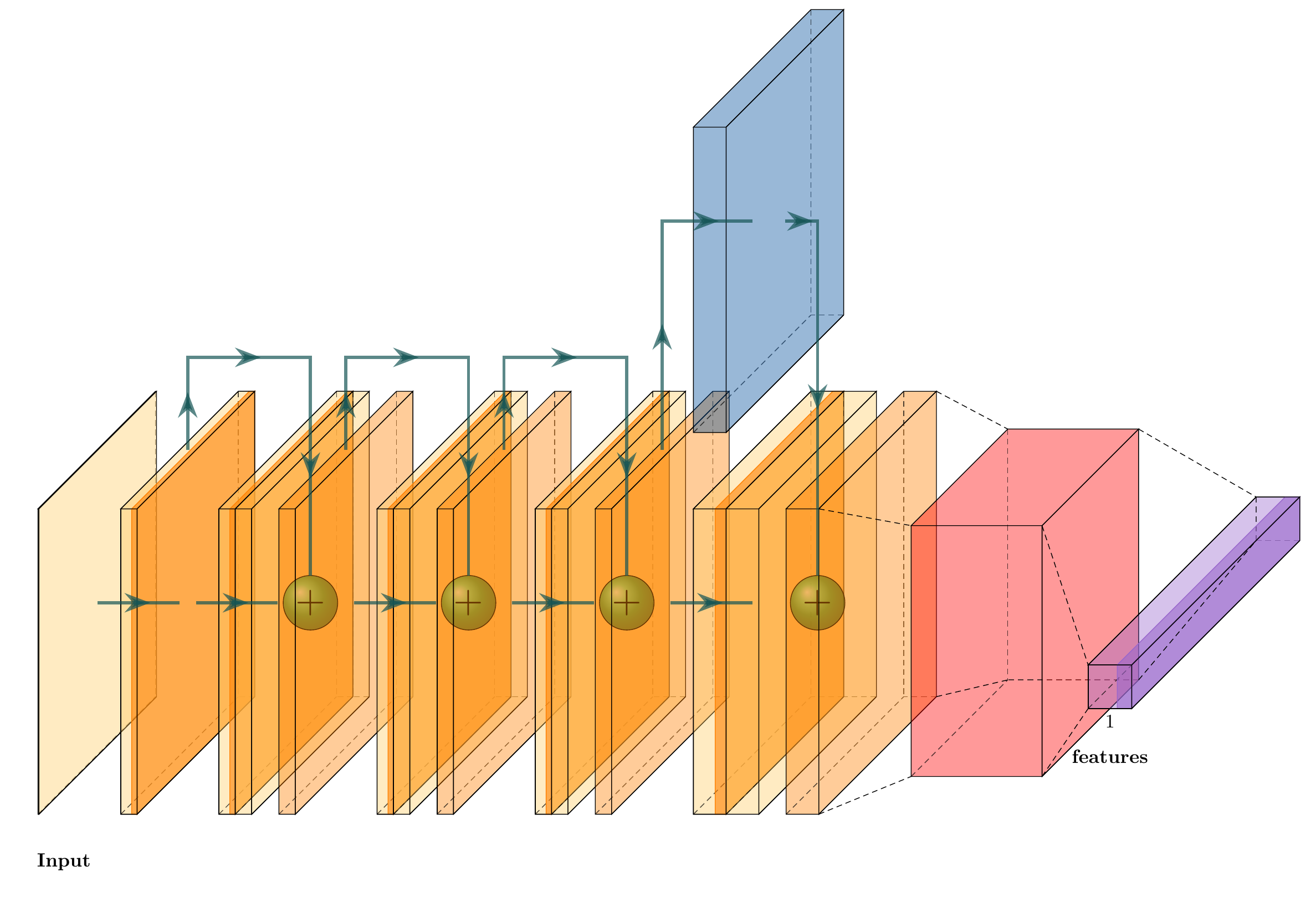}};
\node[rectc](image2) at (12,4) {\includegraphics[scale=0.3]{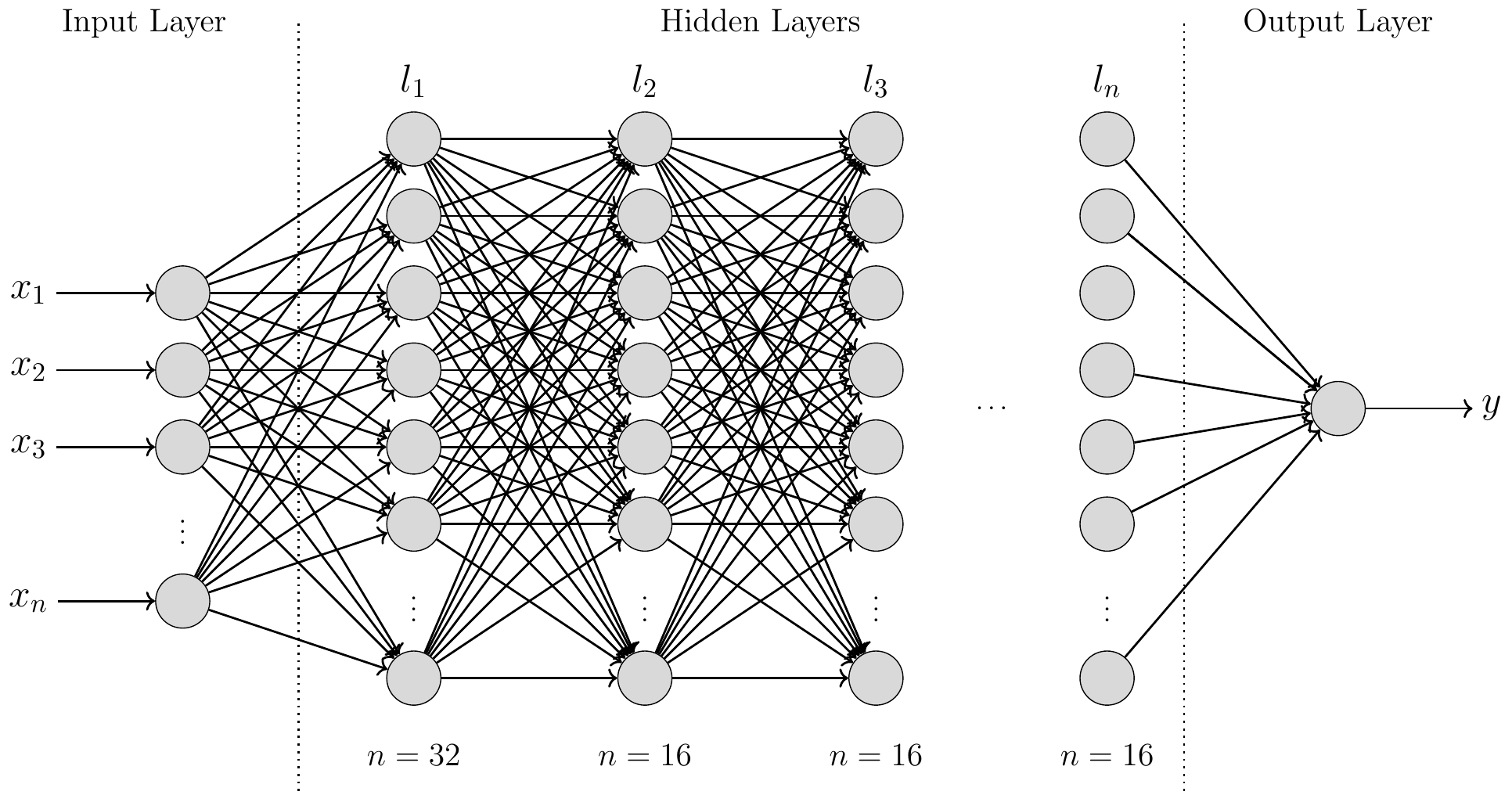}};

\draw[->, thick]  (4,7) -- (3,6);
\draw[->, thick]  (4,7) -- (5,6);
\draw[->, thick]  (12,7) -- (11,6);
\draw[->, thick]  (12,7) -- (13,6);

\draw[->, thick]  (input) -- (feature);
\draw[->, thick]  (feature) -- (processing);
\draw[->, thick]  (processing) -- (Regression);
\draw[->, thick]  (Regression) -- (out);

\end{tikzpicture}}
 \caption{Functional Diagram of Proposed Machine Learning Pipeline.}
 \label{fig:ML_Pipeline}
\end{figure}

The proposed end-to-end machine learning pipeline aimed to relate the visual blockage with hydraulic blockage. It consisted of three main stages (a) visual feature extraction (b) data pre-processing (c) ANN regression. The proposed pipeline was designed to take an image of culvert, extract visual features using deep learning model, pre-process the extracted features and feed into the ANN regression model to predict the hydraulic blockage. Figure \ref{fig:ML_Pipeline} shows the functional block diagram of the proposed machine learning pipeline. 

\subsection{Deep Visual Feature Extraction}

As a fist step in the pipeline, image of culvert is processed through a deep CNN model (e.g., MobileNet, ResNet50, EfficientNetB3) to extract the deep visual features. For this study, three CNN models are compared to assess the impact of number of visual features extracted and fundamental principle by which the visual features are extracted. All the CNN models were used with ImageNet \cite{deng2009imagenet} pre-trained weights and as a feature extractor by removing the top layers. Brief theoretical background of each CNN model is presented as follows. 

\subsubsection{MobileNet}

MobileNets are proposed by Howard et al. \cite{howard2017mobilenets} and are compactly designed fast models for real-world utility. Depthwise separable convolution, a form of factorize convolution \cite{sifre2014rigid} makes them computationally cheaper deep networks. Accuracy and latency of network is controlled by two hyperparameters (i.e., width multiplier, resolution multiplier) to help in building model suitable for a custom problem. Depthwise convolution can be expressed mathematically as in Equation \ref{Equation1}. 

\begin{align}\label{Equation1}
\hat{\mathbf{G}}_{k,l,m}=\sum_{i,j}\hat{\mathbf{K}}_{i,j,m}\cdot \mathbf{F}_{k+i-1, l+j-1, m}
\end{align}

where $\hat{\mathbf{K}}$ denotes the depthwise kernel, $\mathbf{F}$ denotes the input channel and $\hat{\mathbf{G}}$ denotes the filtered output feature map. 

\subsubsection{ResNet50}

Deep residual network was proposed by He et al. \cite{he2016deep} towards improving the training of extremely deep networks by introducing the idea of reformulating layers as learning residual functions instead of unreferenced functions. From empirical results, ResNet proved easier to optimize and improved their accuracy with increased depth of network. In other words, ResNet allow the network layers to fit residual mapping instead of fitting for each layer. If $H(x)$ represents the mapping to be fit by layers of network with input $(x)$, the residual learning is based on the hypothesis that if certain number of layers can asymptotically approximate the complicated function, they can also approximate the residual function $(F(x):= H(x)-x)$.

\subsubsection{EfficientNetB3}

EfficientNet is proposed by Tan and Le \cite{tan2019efficientnet} as an accurate and efficient family of ConvNets based on scaled up version of baseline Neural Architecture Search (NAS) model. Idea of using a simple compound coefficient to uniformly scale the model in all dimensions (i.e., depth, width, resolution) is implemented in developing EfficientNets. Scaling up ConvNets by balancing all dimensions using a constant ratio resulted in better accuracy of models. Based on this idea, if it is intended to use $2^{n}$ times more computational power, model can be scaled up in depth by $a^{n}$, in width by $b^{n}$ and in resolution by $c^{n}$, where $a,b,c$ represent constants. 

\subsection{Data Pre-Processing}

At second step of pipeline, extracted visual features were transformed before feeding to regression model for improved performance. For this study, Standard Scalar transformation was applied which transforms the data with distribution having 0 mean and 1 standard deviation. Given a sample $x$, standard scalar transformation score $z$ can be determined as 

\begin{equation}
    z=\frac{x-\mu}{\sigma}
\end{equation}

where $\mu$ represents the mean and $\sigma$ represents the standard deviation. In literature, it has been reported that standard scalar transformation improves the performance of regression model in comparison to no transformation applied. 

\subsection{ANN Regression}
At the final stage of proposed pipeline, processed visual features were feed into ANN regression model to predict corresponding hydraulic blockage.
ANNs are machine learning models inspired by the biological functionality of animal brain and are layer based deep architecture. ANN consists of nodes, layers and connections. Each node in the network is the representation of a neuron and apply transformation to input by non-linear activation and transmits it to other neurons in the network. Layers of ANN consists of number of nodes and designed to preform specific transformation to input. Furthermore, each layer is characterized by weights which are updated during the training process to optimize the desired performance of layer \cite{abraham2005artificial, mehrotra1997elements, krogh2008artificial, basheer2000artificial}. A neuron $k$ in layer $L+1$ takes $x_{i}^{L}$ as input and transforms it by applying non-linear activation into $x_{k}^{L+1}$. Processing of a single neuron in the network can be mathematically expressed as given in Equation \ref{eqn:ANN}. 

\begin{equation}\label{eqn:ANN}
x_{k}^{L+1}=f\left(\sum_{i} w_{ik}^{L}x_{i}^{L}+w_{bk}^{L}  \right)    
\end{equation}

where $w_{ik}^{L}$ represents the layer $L$ weights, $w_{bk}^{L}$ represents the bias term of neuron $k$ and $f$ represents the non-linear activation function.

\section{Experimental Design}

Experiments were performed with three common deep learning models i.e., MobileNet, ResNet50, EfficientNetB3 as feature extractor to compare the performance and decide best among three to be used in final pipeline. All the CNN models were pre-trained on ImageNet dataset and were used as feature extractor by removing the top layers. Each CNN model resulted in different number of visual features (i.e., MobileNet=50176, ResNet50=100352, EfficientNetB3=153600), therefore, three variants of ANN in terms of number of hidden layers were used to optimize the training. ANN1 was used with MobileNet features, ANN2 was used with ResNet50 features and ANN3 was used with EfficientNetB3 features, respectively. Depth of hidden layers was decided based on trial and error process with the criteria that increasing number of hidden layers do not improve the performance anymore. Table \ref{tab:ANN_Variants} presents the information about the three variants of ANN. 

\begin{table}[]
    \centering
    \caption{ANN Regression Model Variants.}
    \label{tab:ANN_Variants}
    \scalebox{1}{\begin{tabular}{ccccc}
    \toprule
    ~ & \textbf{\# of hidden} & \multirow{2}{*}{\textbf{\# of nodes}} & \textbf{\# of input} & \textbf{\# of trainable} \\
    ~ & \textbf{layers} & ~ & \textbf{features} & \textbf{parameters} \\
    \toprule
    ANN1 & 5 & [32,16,16,16,16] & 50176 & 1607089 \\
    ANN2 & 8 & [32, 16, ... , 16] & 100352 & 3213505 \\
    ANN3 & 10 & [32, 16, ... , 16] & 153600 & 4917985 \\
    \bottomrule
    \end{tabular}}
\end{table}

All the ANN models were trained for 500 epochs with Adam optimizer and constant learning rate of 0.001. Standard 60:20:20 split of dataset was used for training, validation and testing, respectively. Furthermore, Mean Absolute Percentage Error was used as loss metrics during the training process. Models were trained using Nvidia GeForce RTX 2060 Graphical Processing Unit (GPU) with 6GB memory and 14 Gbps memory speed.

Performance of regression models was assessed over unseen test data using standard evaluation metrics including Mean Squared Log Error (MSLE), Mean Squared Error (MSE), Mean Absolute Error (MAE), and $R^{2}$ score. Furthermore, overall machine learning pipeline was assessed for processing times from hardware implementation perspective. 

\section{Results and Discussions}

This section presents the results in terms of regression models numerical results summary, scatter plots, and prediction plots. In addition, overall machine learning pipeline has been assessed for processing times. 

Table \ref{tab:results} presents the summary of recorded quantitative results for the implemented regression models. From the Table \ref{tab:results}, it can be observed that ANN1 model with deep visual features extracted by MobileNet CNN produced the best results with $R^{2}$ of 0.7855. Interestingly, it has been observed that with increase in number of deep visual features, performance of ANN regression degraded. This may be attributed to presence of large number of irrelevant and uncorrelated features for the ResNet50 and EfficientNetB3 cases. 

\begin{table}[H]
    \centering
    \caption{Summary of Empirical Results for Implemented ANN Regression Models}
    \label{tab:results}
    \begin{tabular}{lcccc}
    \toprule
    ~ & \textbf{MSLE} & \textbf{MSE} & \textbf{MAE} & $\mathbf{R^2}$ \\
    \toprule
ANN1 (with MobileNet features)	& 0.10	& 53.54	& 4.67	& 0.7855 \\
ANN2 (with ResNet50 features)	& 0.1522	& 86.92	& 5.21	& 0.6517 \\
ANN3 (with EfficientNetB3 features)	& 0.4903	& 103.75	& 7.03	& 0.5843 \\
\bottomrule
    \end{tabular}
\end{table}

\begin{figure}[H]
\centering
\subfigure[ANN1]{\includegraphics[scale=0.53]{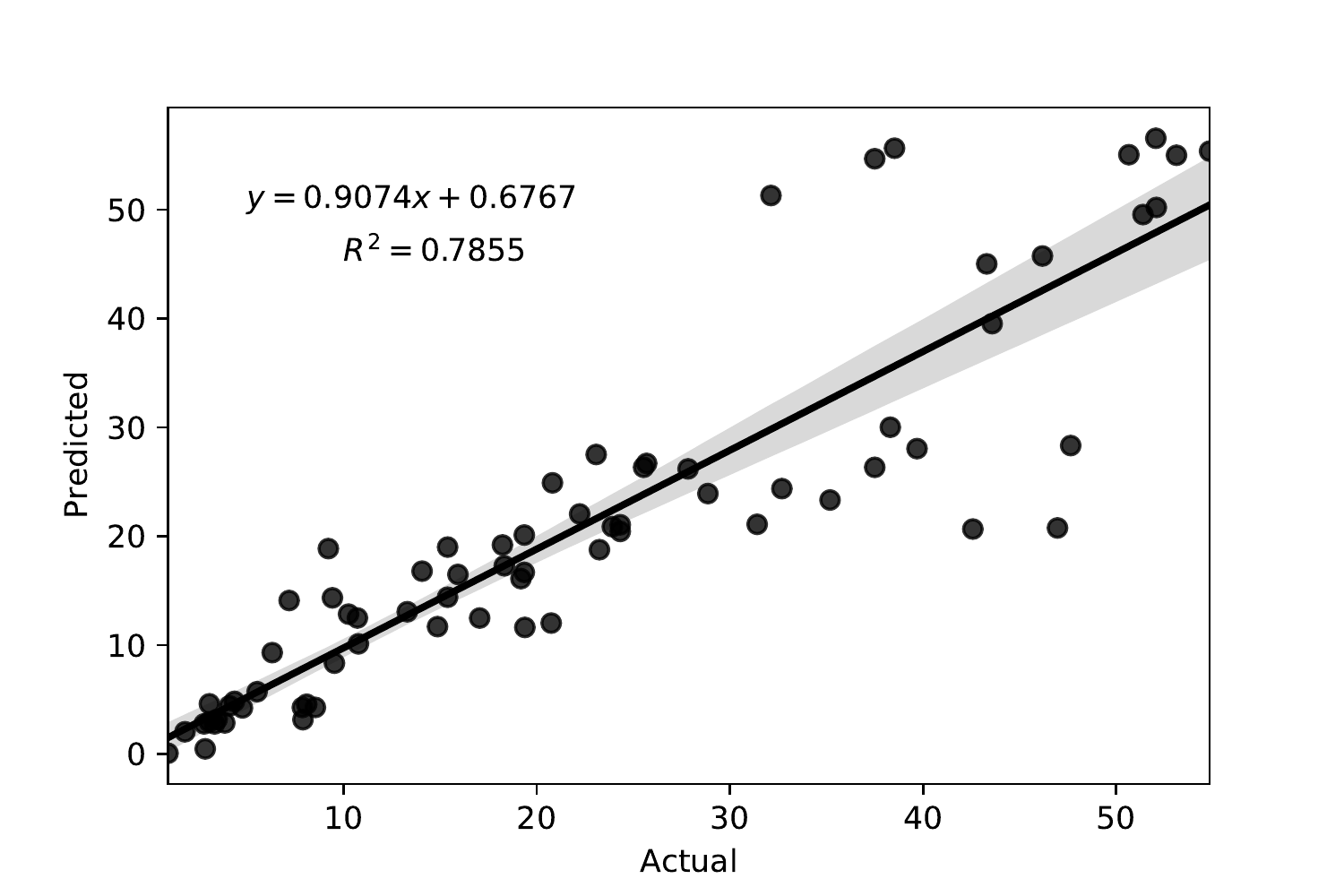}}
\subfigure[ANN2]{\includegraphics[scale=0.53]{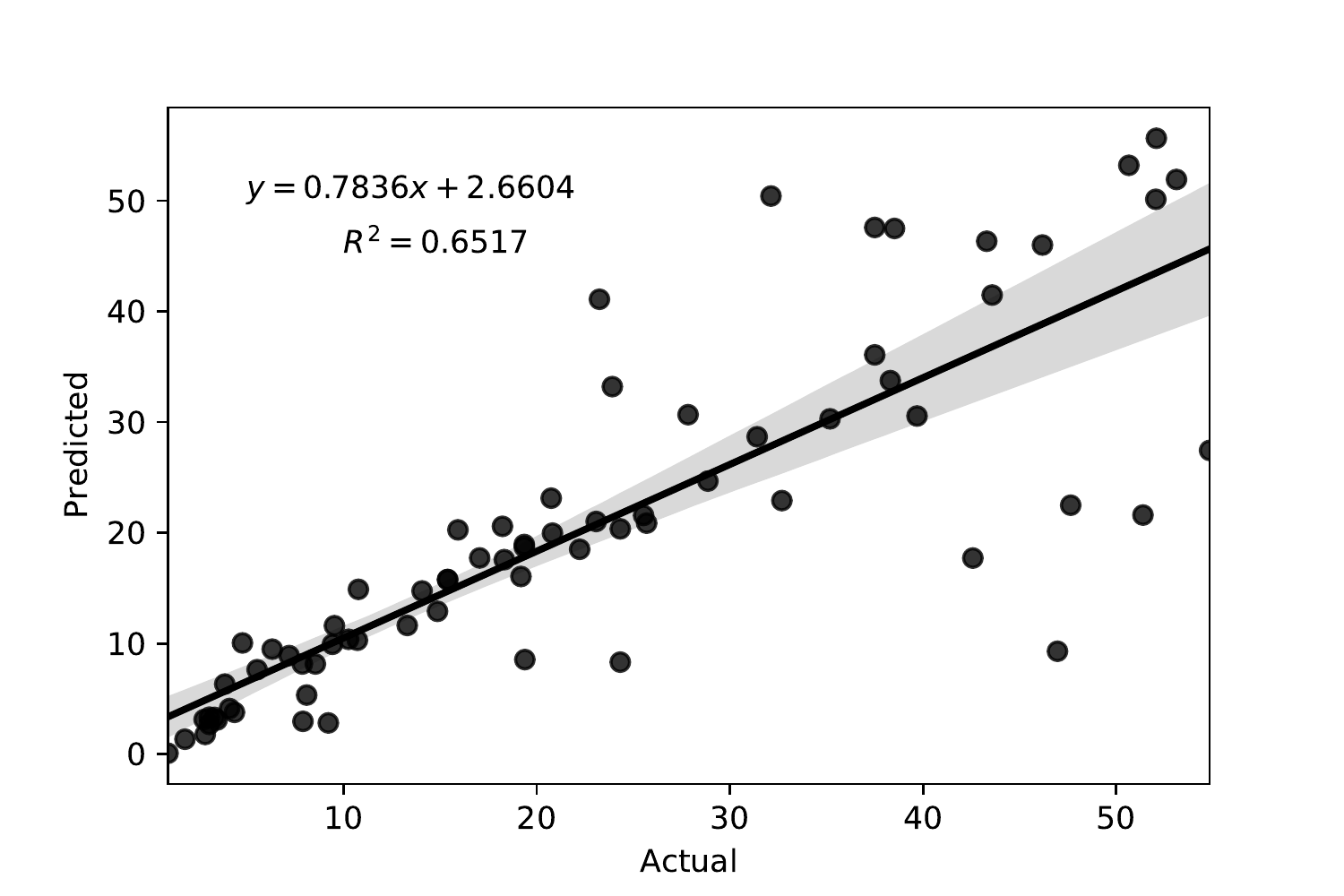}}\\
\subfigure[ANN3]{\includegraphics[scale=0.53]{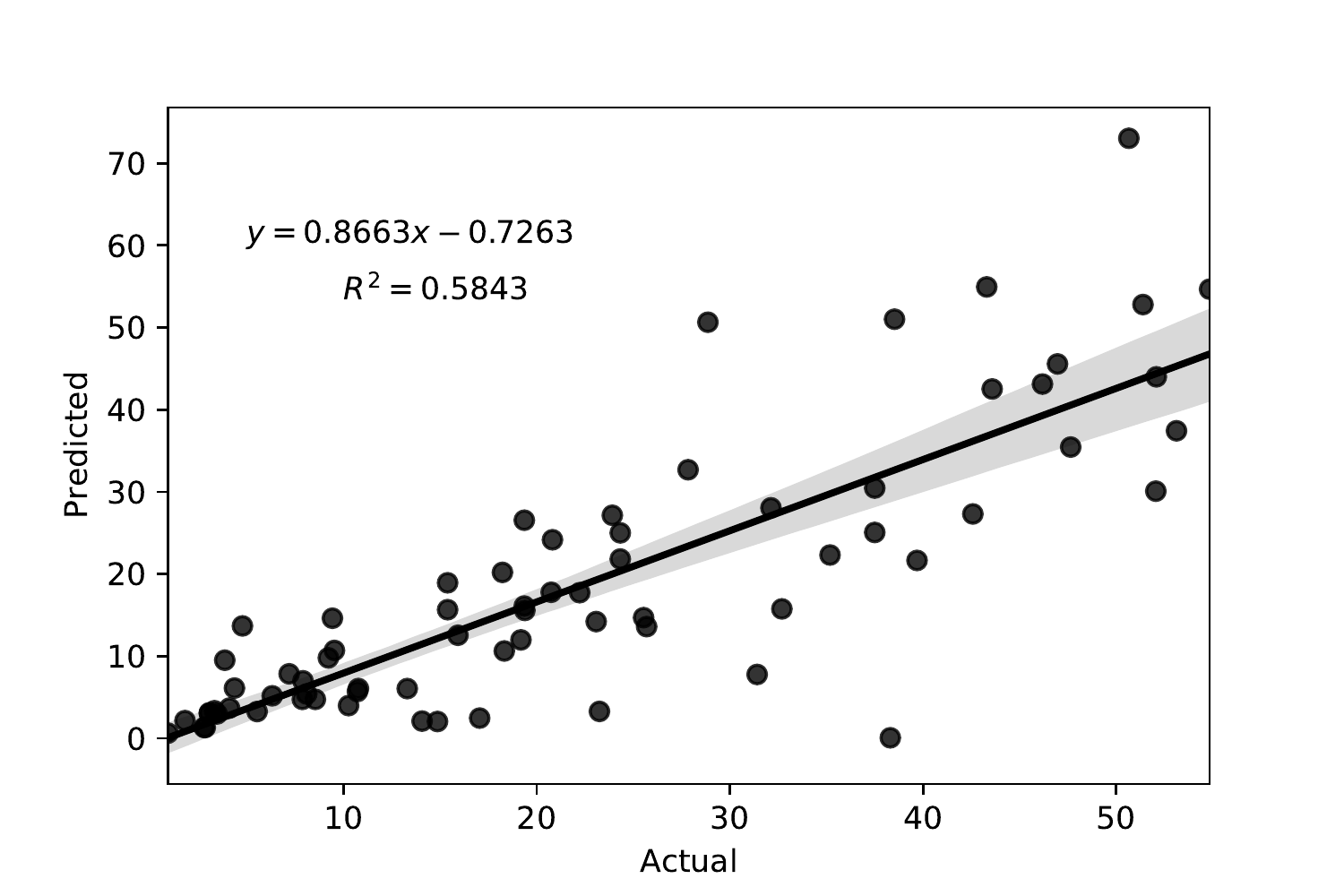}}
\caption{Scatter Plots for Implemented Regression Models to Predict Hydraulic Blockage}
\label{Scatter}
\end{figure}

\begin{figure}[H]
\centering
\subfigure[ANN1]{\includegraphics[scale=0.53]{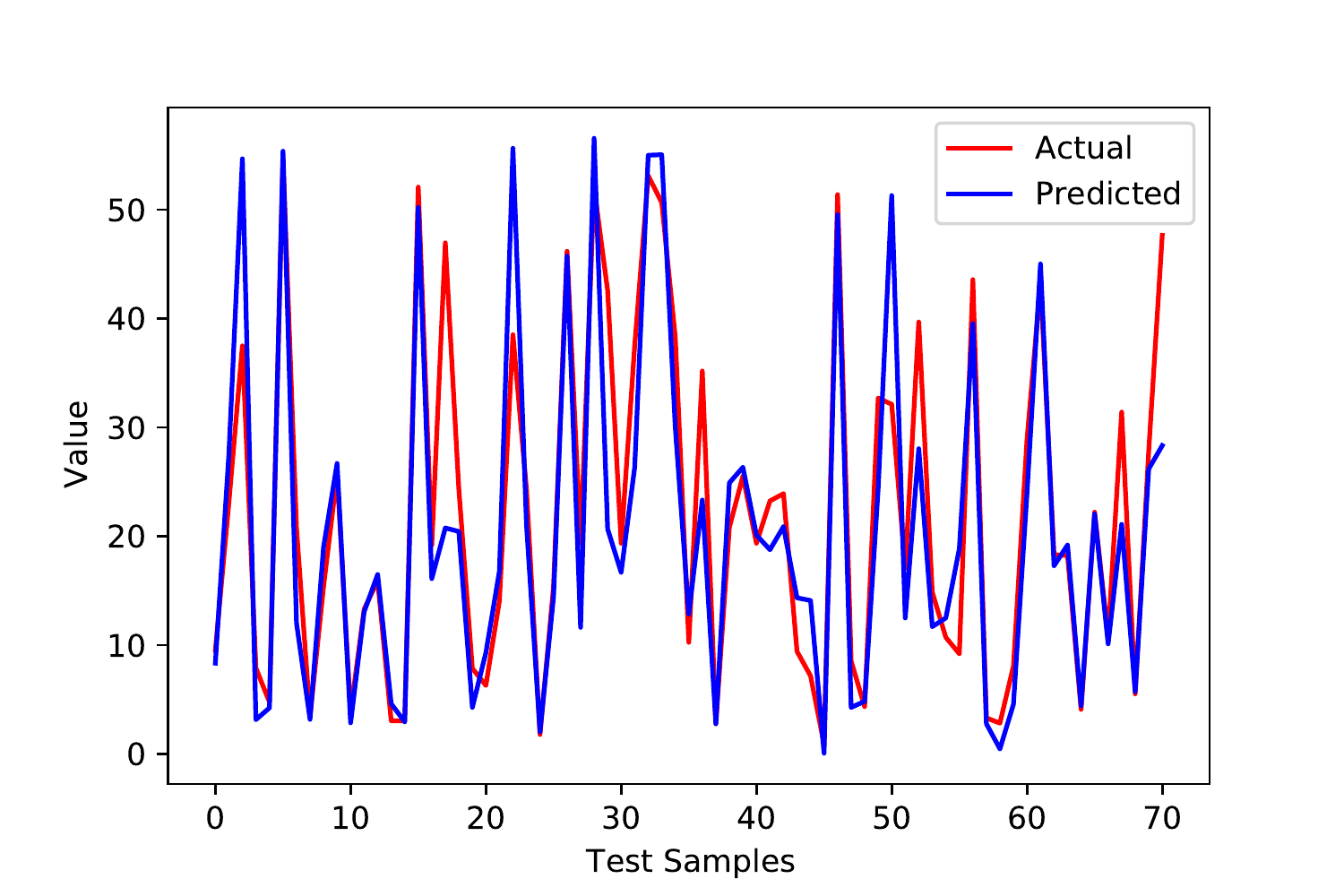}}
\subfigure[ANN2]{\includegraphics[scale=0.53]{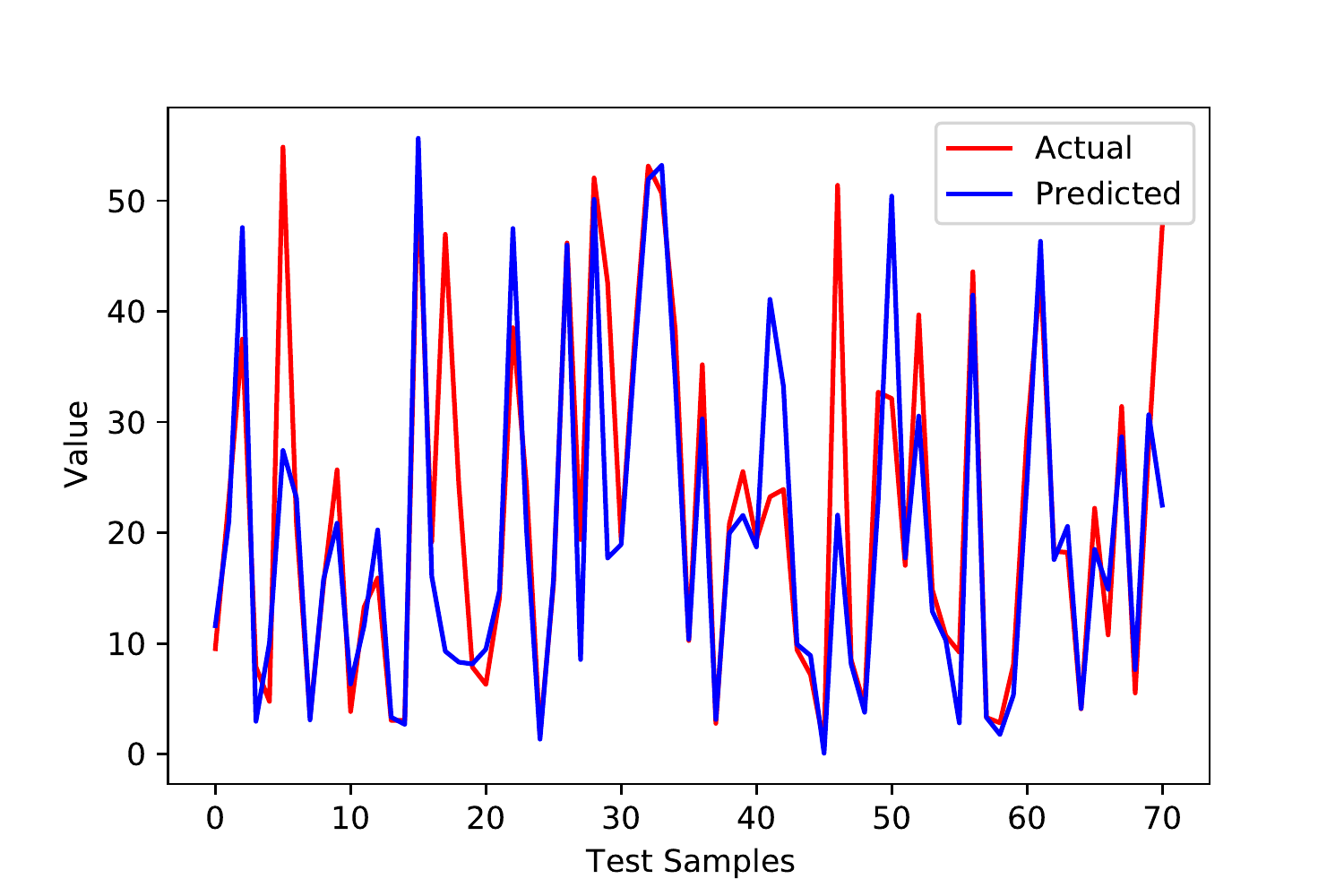}}\\
\subfigure[ANN3]{\includegraphics[scale=0.53]{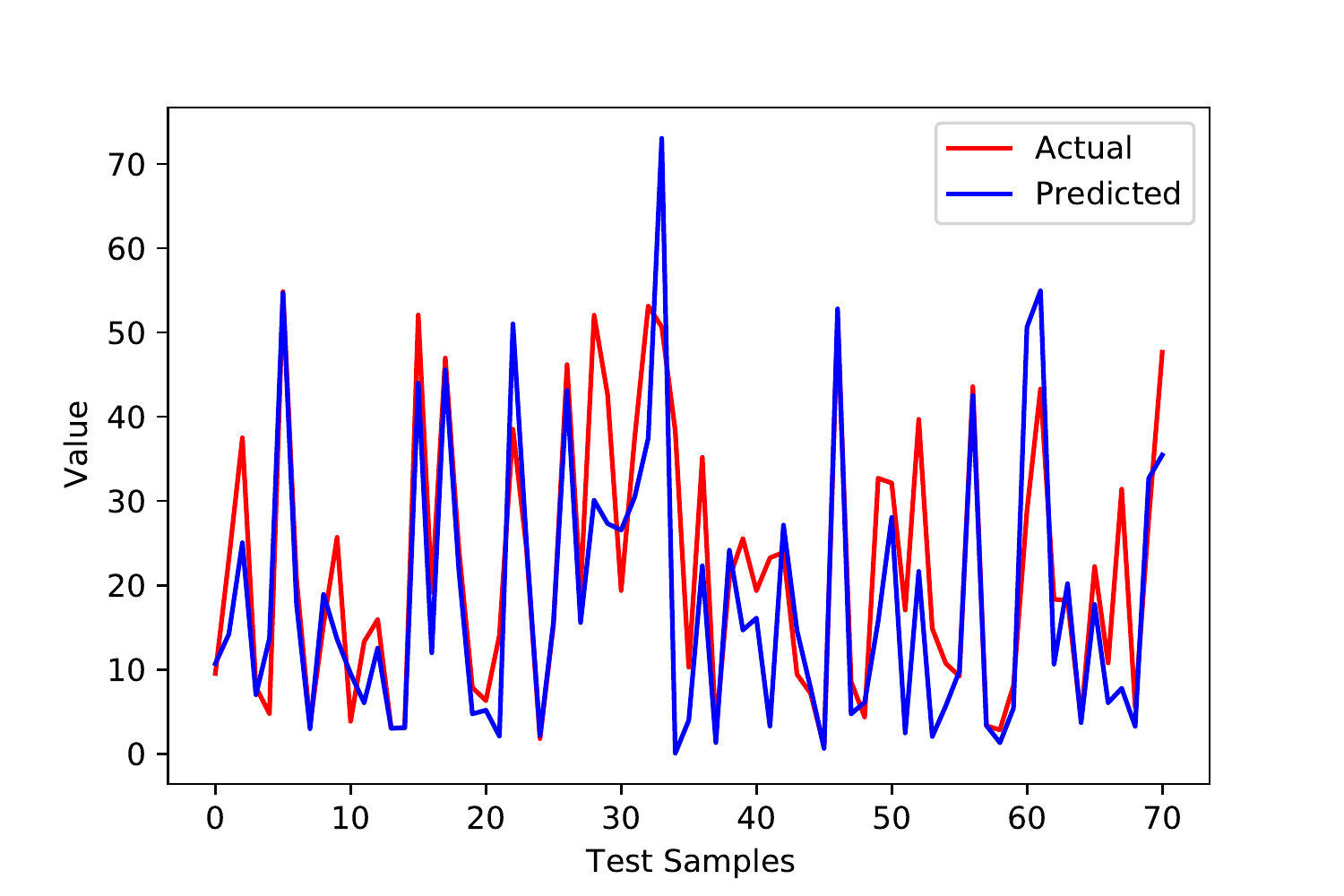}}
\caption{Actual vs Predicted Plots for Implemented Regression Models.}
\label{Prediction}
\end{figure}

Figure \ref{Scatter} shows the scatter plots for each ANN regression model. From the plots, it is evident that ANN1 with MobileNet extracted deep visual features produced the best fit on test data. Figure \ref{Prediction} shows the actual vs predicted plots for all three ANN models to demonstrate how well each model was able to track the actual value. ANN1 model was observed to best track the actual values, however, over-prediction can be observed at majority data instances. In all three cases, over-prediction was more dominant in comparison to under-prediction. 

Maximum $R^{2}$ score of 0.7855 for MobileNet and positive for all three cases indicated the presence of correlation between visual features and hydraulic blockage which suggested that both visual blockage and hydraulic blockage can be interrelated. However, it is important to mention that dataset used for this investigation was recorded with same background and lighting conditions with only variations in culvert type, debris and water levels. This suggests that, for real-world application, as part of calibration process, camera should focus only on culvert region avoiding any vegetative background otherwise performance may significantly degrade given the visual similarity between vegetative background and vegetative debris material causing blockage. 

The proposed machine learning pipeline was assessed in terms of response times for relative comparison of models towards hardware implementation. Table \ref{tab:Response_Time} shows the processing times for each component of proposed pipeline. It can be observed that ANN1 with MobileNet features was the fastest in predicting the hydraulic blockage while ANN3 with EfficientNetB3 features was slowest. Degraded response time may be attributed to increased number of features and increased depth of ANN model. Reported processing times are not the true measure of cutting edge hardware performance rather are for relative comparison purposes. However, technological advancements in the computing hardware and availability of cutting edge hardware (e.g., Nvidia Jetson Nano, Nvidia Jetson TX2) has made it possible to implement machine learning and computer vision algorithms for real-world problems (e.g., tracking of pedestrian traffic \cite{barthelemy2019edge}, tracking of wildlife \cite{arshad2020my}). 

\begin{table}[H]
    \centering
    \caption{Comparison of Machine Learning Pipeline Components for Response Times.}
    \label{tab:Response_Time}
    \scalebox{0.75}{\begin{tabular}{ccccccc}
    \toprule
    ~ & \multirow{2}{*}{\textbf{Image Loading}} & \textbf{CNN Model} & \multirow{2}{*}{\textbf{Feature Extraction}} & \multirow{2}{*}{\textbf{Scalar Transform}} & \textbf{ANN Model} & \multirow{2}{*}{\textbf{Prediction}} \\
    ~ & ~ & \textbf{Loading} & ~ & ~ & \textbf{Loading} & ~ \\
    \toprule
ANN1 & 0.0966 sec	& 0.7545 sec	& 0.1122 sec	& 0.0183 sec	& 0.2172 sec	& 0.0502 sec \\
ANN2 & 0.1030 sec	& 2.1470 sec	& 0.2441 sec	& 0.0175 sec	& 0.2980 sec	& 0.0637 sec\\
ANN3 & 0.1039 sec	& 2.9691 sec	& 0.2976 sec	& 0.0193 sec	& 0.3967 sec	& 0.0751 sec \\
    \bottomrule
    \end{tabular}}
\end{table}

\section{Conclusion}

An end-to-end machine learning pipeline has been successfully implemented to relate visual blockage with hydraulic blockage by proposing the regression on deep visual features. Three CNN models (i.e., MobileNet, ResNet50, EfficientNetB3) were compared for number of extracted visual features and fundamental principle by which features are extracted. Artificial Neural Network (ANN) regression model was implemented on deep visual features to predict the hydraulic blockage. From results, it has been reported that ANN with MobileNet extracted visual features (i.e., ANN1) achieved the best regression performance with $R^{2}$ of 0.7855. Regression performance was observed to be degraded with increase in number of extracted visual features. The degraded performance may be attributed to presence of increased number of irrelevant and uncorrelated features in case of ResNet50 and EfficientNetB3. Positive $R^{2}$ score for all cases with maximum for MobileNet indicated the presence of correlation between visual features and hydraulic blockage and suggested that visual blockage and hydraulic blockage can be interrelated. Performance of regression models is expected to be degraded significantly for the cases where image contain background with similar visual appearance to the debris material blocking the culvert. With the availability of hydraulic and visual data from real-world events, extension of proposed approach for real-world application is a potential future direction for presented research.  

\section*{Acknowledgment}
I would like to thank the Wollongong City Council (WCC) for funding this investigation. This research was funded by the Smart Cities and Suburb Program (Round Two) of the Australian Government, grant number SCS69244. Further, I would like to thank the Higher Education Commission (HEC) of Pakistan and the University of Wollongong (UOW) for funding my PhD studies.

\bibliographystyle{IEEEtran}
\bibliography{IEEEabrv,references}

\begin{thebibliography}{10}
\providecommand{\url}[1]{#1}
\csname url@samestyle\endcsname
\providecommand{\newblock}{\relax}
\providecommand{\bibinfo}[2]{#2}
\providecommand{\BIBentrySTDinterwordspacing}{\spaceskip=0pt\relax}
\providecommand{\BIBentryALTinterwordstretchfactor}{4}
\providecommand{\BIBentryALTinterwordspacing}{\spaceskip=\fontdimen2\font plus
\BIBentryALTinterwordstretchfactor\fontdimen3\font minus
  \fontdimen4\font\relax}
\providecommand{\BIBforeignlanguage}[2]{{%
\expandafter\ifx\csname l@#1\endcsname\relax
\typeout{** WARNING: IEEEtran.bst: No hyphenation pattern has been}%
\typeout{** loaded for the language `#1'. Using the pattern for}%
\typeout{** the default language instead.}%
\else
\language=\csname l@#1\endcsname
\fi
#2}}
\providecommand{\BIBdecl}{\relax}
\BIBdecl

\bibitem{french2015culvert}
R.~French and M.~Jones, ``Culvert blockages in two australian flood events and
  implications for design,'' \emph{Australasian Journal of Water Resources},
  vol.~19, no.~2, pp. 134--142, 2015.

\bibitem{french2018design}
------, ``Design for culvert blockage: the arr 2016 guidelines,''
  \emph{Australasian Journal of Water Resources}, vol.~22, no.~1, pp. 84--87,
  2018.

\bibitem{BarthelmessMechanism}
A.~Barthelmess and E.~Rigby, ``Culvert blockage mechanisms and their impact on
  flood behaviour,'' in \emph{Proceedings of the 34th World Congress of the
  International Association for Hydro- Environment Research and
  Engineering}.\hskip 1em plus 0.5em minus 0.4em\relax Barton, ACT: Engineers
  Australia, 2011, pp. 380--387.

\bibitem{rigby2001impact}
E.~Rigby and P.~Silveri, ``The impact of blockages on flood behaviour in the
  wollongong storm of august 1998,'' in \emph{6th Conference on Hydraulics in
  Civil Engineering: The State of Hydraulics}.\hskip 1em plus 0.5em minus
  0.4em\relax Barton, ACT: Engineers Australia, 2001, pp. 107--115.

\bibitem{roso2004prediction}
S.~Roso, M.~Boyd, E.~Rigby, and R.~VanDrie, ``Prediction of increased flooding
  in urban catchments due to debris blockage and flow diversions,'' in
  \emph{Proceedings of NOVATECH}, 2004, pp. 8--13.

\bibitem{rigby2002causes}
E.~Rigby and P.~Silveri, ``Causes and effects of culvert blockage during large
  storms,'' in \emph{Ninth International Conference on Urban Drainage
  (9ICUD)}.\hskip 1em plus 0.5em minus 0.4em\relax Lloyd Center Doubletree
  Hotel, Portland, Oregon, United States: Engineers Australia, September 2002,
  pp. 1--16.

\bibitem{davis2001analysis}
A.~Davis, ``An analysis of the effects of debris caught at various points of
  major catchments during wollongong’s august 1998 storm event,''
  \emph{Bachelor of Engineering Thesis, University of Wollongong}, 2001.

\bibitem{iqbalcomputer}
U.~Iqbal, P.~Perez, W.~Li, and J.~Barthelemy, ``How computer vision can
  facilitate flood management: A systematic review,'' \emph{International
  Journal of Disaster Risk Reduction}, vol.~53, p. 102030, 2021.

\bibitem{kramer2015physical}
M.~Kramer, W.~Peirson, R.~French, and G.~Smith, ``A physical model study of
  culvert blockage by large urban debris,'' \emph{Australasian Journal of Water
  Resources}, vol.~19, no.~2, pp. 127--133, 2015.

\bibitem{blanc2013analysis}
J.~Blanc, ``An analysis of the impact of trash screen design on debris related
  blockage at culvert inlets,'' Ph.D. dissertation, School of the Built
  Environment, Heriot-Watt University, 2013.

\bibitem{blanc2014analysis}
J.~Blanc, N.~P. Wallerstein, S.~Arthur, and G.~B. Wright, ``Analysis of the
  performance of debris screens at culverts,'' in \emph{Proceedings of the
  Institution of Civil Engineers-Water Management}, vol. 167, no.~4.\hskip 1em
  plus 0.5em minus 0.4em\relax Thomas Telford Ltd, 2014, pp. 219--229.

\bibitem{NSWfloodplain}
NSW, \emph{Floodplain Development Manual}, New South Wales Government, Sydney,
  Australia, 2005.

\bibitem{jones1991queensland}
N.~Jones and C.~Lawson, ``The queensland urban drainage manual,'' \emph{Local
  Government Engineers Association of Queensland Journal}, vol.~9, no. 4th
  quarter, 1991.

\bibitem{ollett2017australian}
P.~Ollett, B.~Syme, and P.~Ryan, ``Australian rainfall and runoff guidance on
  blockage of hydraulic structures: numerical implementation and three case
  studies,'' \emph{Journal of Hydrology (New Zealand)}, vol.~56, no.~2, pp.
  109--122, 2017.

\bibitem{ARR_Report}
W.~Weeks, G.~Witheridge, E.~Rigby, A.~Barthelmess, and G.~O‘Loughlin,
  ``Project 11: Blockage of hydraulic structures,'' Engineers Australia, Water
  Engineering, 11 National Circuit Barton ACT 2600, Tech. Rep. P11/S2/021,
  February 2013.

\bibitem{ball2016australian}
J.~Ball, M.~Babister, R.~Nathan, P.~Weinmann, W.~Weeks, M.~Retallick, and
  I.~Testoni, ``Australian rainfall and runoff-a guide to flood estimation,''
  2016.

\bibitem{CondouitBlockagePolicy}
R.~H. Jones, W.~Weeks, and M.~Babister, \emph{Review of Conduit Blockage Policy
  Summary Report}.\hskip 1em plus 0.5em minus 0.4em\relax 160 Clarence Street
  Sydney, NSW, 2000: WMA Water, June 2016.

\bibitem{Barthelmess2009Quantification}
A.~Barthelmess and E.~Rigby, ``Quantification of debris potential and the
  evolution of a regional culvert blockage model,'' in \emph{32nd Hydrology and
  Water Resources Symposium}.\hskip 1em plus 0.5em minus 0.4em\relax Barton,
  ACT: Engineers Australia, 2009, pp. 218--229.

\bibitem{french2012non}
R.~French, E.~Rigby, and A.~Barthelmess, ``The non-impact of debris blockages
  on the august 1998 wollongong flooding,'' \emph{Australasian Journal of Water
  Resources}, vol.~15, no.~2, pp. 161--169, 2012.

\bibitem{deng2009imagenet}
J.~Deng, W.~Dong, R.~Socher, L.-J. Li, K.~Li, and L.~Fei-Fei, ``Imagenet: A
  large-scale hierarchical image database,'' in \emph{Proceedings of the IEEE
  conference on computer vision and pattern recognition}.\hskip 1em plus 0.5em
  minus 0.4em\relax Miami, FL, USA: IEEE, June 2009, pp. 248--255.

\bibitem{howard2017mobilenets}
A.~G. Howard, M.~Zhu, B.~Chen, D.~Kalenichenko, W.~Wang, T.~Weyand,
  M.~Andreetto, and H.~Adam, ``Mobilenets: Efficient convolutional neural
  networks for mobile vision applications,'' \emph{arXiv preprint
  arXiv:1704.04861}, 2017.

\bibitem{sifre2014rigid}
L.~Sifre and S.~Mallat, ``Rigid-motion scattering for image classification,''
  Ph.D. dissertation, Ecole Polytechnique, October 2014.

\bibitem{he2016deep}
K.~He, X.~Zhang, S.~Ren, and J.~Sun, ``Deep residual learning for image
  recognition,'' in \emph{Proceedings of the IEEE conference on computer vision
  and pattern recognition}.\hskip 1em plus 0.5em minus 0.4em\relax Las Vegas,
  NV, USA: IEEE, June 2016, pp. 770--778.

\bibitem{tan2019efficientnet}
M.~Tan and Q.~V. Le, ``Efficientnet: Rethinking model scaling for convolutional
  neural networks,'' \emph{arXiv preprint arXiv:1905.11946}, 2019.

\bibitem{abraham2005artificial}
A.~Abraham, ``Artificial neural networks,'' \emph{Handbook of measuring system
  design}, 2005.

\bibitem{mehrotra1997elements}
K.~Mehrotra, C.~K. Mohan, and S.~Ranka, \emph{Elements of artificial neural
  networks}.\hskip 1em plus 0.5em minus 0.4em\relax MIT press, 1997.

\bibitem{krogh2008artificial}
A.~Krogh, ``What are artificial neural networks?'' \emph{Nature biotechnology},
  vol.~26, no.~2, pp. 195--197, 2008.

\bibitem{basheer2000artificial}
I.~A. Basheer and M.~Hajmeer, ``Artificial neural networks: fundamentals,
  computing, design, and application,'' \emph{Journal of microbiological
  methods}, vol.~43, no.~1, pp. 3--31, 2000.

\bibitem{barthelemy2019edge}
J.~Barth{\'e}lemy, N.~Verstaevel, H.~Forehead, and P.~Perez, ``Edge-computing
  video analytics for real-time traffic monitoring in a smart city,''
  \emph{Sensors}, vol.~19, no.~9, p. 2048, 2019.

\bibitem{arshad2020my}
B.~Arshad, J.~Barthelemy, E.~Pilton, and P.~Perez, ``Where is my deer?-wildlife
  tracking and counting via edge computing and deep learning,'' in \emph{2020
  IEEE Sensors}.\hskip 1em plus 0.5em minus 0.4em\relax IEEE, 2020, pp. 1--4.

\end{thebibliography}

\end{document}